%% file: main.tex
\def\inreview{1}

\if\inreview1
\documentclass[authoryear,review]{elsarticle}
\else
\documentclass[preprint,12pt,authoryear]{elsarticle}
\fi

\usepackage{lipsum}
\usepackage[utf8]{inputenc}
\usepackage[T1]{fontenc}
\usepackage[english]{babel}

\usepackage{tabularx}

\usepackage{xcolor}

\usepackage{xspace}

\usepackage{amsmath}
\usepackage{amssymb}
\usepackage{amsfonts}
\usepackage{amsthm}

\usepackage{algorithm}
\usepackage{algpseudocode}

\usepackage{hyperref}

\usepackage{listings}
\usepackage{float}

\usepackage{placeins}
\let\oldsection\section
\renewcommand{\section}{\FloatBarrier\oldsection}
\let\oldsubsection\subsection
\renewcommand{\subsection}{\FloatBarrier\oldsubsection}

\usepackage{ulem}

\usepackage{orcidlink}
\usepackage{makecell}
\usepackage{booktabs}

\definecolor{mygreen}{rgb}{0,0.6,0}
\definecolor{mygray}{rgb}{0.5,0.5,0.5}
\definecolor{mymauve}{rgb}{0.58,0,0.82}
\definecolor{azure}{rgb}{0.0, 0.5, 1.0}

\newcommand{\skc}{{\color{black}{\it Scikit-Criteria}}\xspace}

\newcommand{\mcda}{{\color{black}{\textsc{mcda}}}\xspace}
\newcommand{\mcdm}{{\color{black}{\textsc{mcdm}}}\xspace}

\usepackage{xcolor}
\usepackage{tcolorbox}

 \journal{European Journal of Operational Research}

\begin{document}

% The title page
% \if\inreview1
% \input{ES&C/title_page}
% \fi

\begin{frontmatter}



\title{Closing a 17-Year Gap: Algorithmic Detection and Empirical Prevalence of Rank Reversal in Multi-Criteria Decision Analysis}

\author[conae,conicet,famaf]{Cabral, Juan Bautista \orcidlink{0000-0002-7351-0680} \corref{cor1}}
\ead{jbcabral@unc.edu.ar}
\author[ifeg, famaf]{Giarda, Gonzalo \orcidlink{0009-0005-0598-0372}} \ead{gonzalo.giarda@unc.edu.ar}
\author[ipqa,famaf]{Gimenez Irusta, Diego Nicolás \orcidlink{0009-0001-5536-1634}}\ead{diego.gimenez@unc.edu.ar}
\author[conae,conicet,famaf]{Pacheco, Paula \orcidlink{0009-0005-0005-1515}}\ead{paulapacheco@unc.edu.ar}
\author[iate,famaf]{Schachner, Alvaro Roy \orcidlink{0009-0008-5460-8720}}\ead{alvaro.schachner@mi.unc.edu.ar}
\author[unrc,conicet, famaf]{Borda, Agustín \orcidlink{0009-0001-8548-2998}} \ead{aborda@dc.exa.unrc.edu.ar}

\affiliation[conae]{organization={Grupo de Innovación y Desarrollo Tecnológico, Gerencia de Vinculación Tecnológica, Centro Espacial Teófilo Tabanera, Comisión Nacional de Actividades Espaciales (GVT-CONAE)}, postcode={5187},
city={Falda del Cañete, Córdoba},
country={Argentina}}

\affiliation[conicet]{organization={Consejo Nacional de Investigaciones Científicas y Técnicas (CONICET)},
                     addressline={Godoy Cruz 2290},
                     postcode={C1425FQB},
                     city={Ciudad Autónoma de Buenos Aires},
                     country={Argentina}}

\affiliation[famaf]{organization={Facultad de Matem{\'a}tica, Astronom{\'\i}a, F{\'\i}sica y Computaci{\'o}n, Universidad Nacional de Córdoba (FAMAF-UNC)},
                   city={Córdoba},
                   postcode={5016},
                   country={Argentina}}

\affiliation[iate]{organization={Instituto de Astronomía Teórica y Experimental, Universidad Nacional de Córdoba - Consejo Nacional de Investigaciones Científicas y Técnicas (IATE, UNC-CONICET)},
                  addressline={Laprida 854},
                  postcode={X5000BGR},
                  city={Córdoba},
                  country={Argentina}}

\affiliation[ipqa]{organization={Instituto de Investigación y Desarrollo en Ingeniería de Procesos y Química Aplicada, Universidad Nacional de Córdoba - Consejo Nacional de Investigaciones Científicas y Técnicas (IPQA, UNC-CONICET)},
                  addressline={Vélez Sársfield 1611},
                  postcode={5016},
                  city={Córdoba},
                  country={Argentina}}

\affiliation[unrc]{organization={Departamento de Computaci\'on, Universidad Nacional de Río Cuarto (UNRC)},
                  city={Río Cuarto},
                  postcode={X5804BYA},
                  country={Argentina}}

\affiliation[ifeg]{organization={Instituto de Física Enrique Gaviola (IFEG, UNC-CONICET)},
                  city={Córdoba},
                  postcode={X5000HUA},
                  country={Argentina}}

\cortext[cor1]{Corresponding author}

\begin{abstract}
Rank Reversal, where the relative order of alternatives changes in ways that violate axioms of rational decision-making, is a well-documented threat to the reliability of Multi-Criteria Decision Analysis (MCDA) methods. \citet{wang2008ranking} proposed three systematic test criteria to detect this phenomenon, but despite more than 700 citations, no validated, open-source implementation has closed the gap between theory and practice, a 17-year absence we trace to the non-trivial algorithmic challenges of operationalizing these tests for real-world pipelines.
We present an algorithmic framework, implemented in the open-source \skc library, that translates \citet{wang2008ranking}'s three criteria into concrete, pipeline-compatible procedures: a controlled degradation strategy with hierarchical tie-breaking and graceful handling of preprocessing filters (RRT1), and a dominance-graph construction with exact condensation and transitive reduction for detecting transitivity violations and recomposition inconsistencies (RRT2/RRT3).
We demonstrate the framework through two case studies: an application to a cryptocurrency evaluation problem \citep{vanheerden2021crypto}, and a large-scale audit of 27 pipeline/dataset combinations reproduced from 20 published \mcdm methods. The audit shows top-alternative stability (RRT1) is nearly universal (96.3\%), but transitivity (RRT2) fails for 14.8\% and recomposition consistency (RRT3), the strictest criterion, fails for nearly half (48.1\%) of published examples—evidence that rank reversal is a pervasive, measurable feature of the current \mcdm literature, not a marginal or adversarial concern.
\end{abstract}


\end{frontmatter}

%\tableofcontents



\section{Introduction}
\label{introduction}

Multi-Criteria Decision Analysis (\mcda) represents a systematic framework for evaluating complex decision problems involving multiple, often conflicting objectives and criteria. The central idea behind \mcda is to integrate multiple evaluation criteria into coherent rankings or selections of alternatives through structured procedures that manage multi-dimensional complexity.
Numerous methods have been developed to systematically compare alternatives while optimizing for domain-specific requirements and preferences \citep{papathanasiou2018topsis}. These techniques have been successfully applied to multiple and diverse domains, including health care and finance \citep{zopounidis2015multiple}, demonstrating their versatility in identifying optimal solutions for varied decision contexts.

{Despite this widespread success, \mcda methods are susceptible to a phenomenon called Rank Reversals, a situation where the order of alternatives changes when the set of alternatives changes. This counterintuitive behavior can violate axioms of rational decision-making theory, such as \textit{Independence of irrelevant alternatives} and \textit{transitivity}. When rank reversals occur, they can undermine the reliability and logical consistency of \mcda methods, potentially leading to suboptimal or contradictory decisions in critical applications. These violations are manifested through several mechanisms. Five types of rank reversals have been identified in the literature \citep{aires2018rank}. Type I occurs when the final rank order of the alternatives changes if an irrelevant alternative is added to (or removed from) the problem. Type II when the indication of the best alternative changes if a non-optimal alternative is replaced by another worse one. Type III appears when the transitivity property is violated if an irrelevant alternative is added to (or removed from) the problem. Type IV appears when the transitivity property is violated through problem decomposition, where rankings of smaller sub-problems conflict with the overall ranking of alternatives. Finally, Type V occurs when the final rank order changes upon removing a non-discriminating criterion, despite such criteria providing no differential information between alternatives.}

How to interpret these violations, however, remains contested. The positivist view \citep{wang2008ranking} treats rank reversals as irregularities that a reliable method should not exhibit, proposing test criteria under which an effective \mcda method is expected to preserve the indication of the best alternative and satisfy transitivity when a decision problem is decomposed into pairwise comparisons. The constructivist perspective \citep{figueira2009} counters that such instability does not invalidate the methods themselves, but rather reflects the poorly determined margins inherent to purely ordinal data; since decision aiding is not a tool for uncovering a pre-existing ``true'' ranking. The pragmatic approach \citep{yang2020} sidesteps this philosophical debate altogether, instead formally defining ranking stability and proving that specific method variants satisfy it by construction, recognizing that users need reliable results regardless of philosophical stance.
These rank reversal problems have been identified in major \mcda approaches, including \textsc{topsis} \citep{papathanasiou2018topsis}, \textsc{electre}-type methods \citep{roy1978electre}, and \textsc{promethee} \citep{brans2005promethee}, demonstrating that this is a fundamental challenge rather than a limitation of specific techniques.

To address these reliability issues, \citet{wang2008ranking} proposed three systematic test criteria designed to detect and quantify rank reversal behavior in \mcda methods. {These criteria provide} a theoretical framework for evaluating method stability by examining: (1) the persistence of optimal alternatives under controlled degradation of suboptimal options, (2) the preservation of transitivity in pairwise decompositions, and (3) the consistency of rankings reconstructed from problem partitions.

{It is important to note that while \citet{wang2008ranking} provided conceptual criteria rather than implementable algorithms, and despite more than 700 citations since 2008, a review of the literature reveals that there are no comprehensive, validated, and open-source computational implementations inspired by these fundamental tests. Section~\ref{section:from-theory} examines why this gap has persisted and details the specific algorithmic challenges involved in closing it.}

To address this {implementation gap}, this work develops an algorithmic framework for rank reversal detection and analysis within \skc, an open-source Python library specifically designed for \mcda \citep{cabral2016scikit}.
\skc provides a unified interface for various \mcda methods, incorporating preprocessing pipelines, result comparison tools, and extensible architectures that facilitate method development and evaluation.
Our contribution is fundamentally methodological and computational rather than theoretical. We translate \citet{wang2008ranking}'s theoretical test criteria into practical, computational, and extensible algorithmic tools that: (1) handle real-world complexities through principled design choices (graceful degradation, hierarchical tie-breaking); (2) integrate seamlessly with modern \mcdm pipelines; (3) provide complete transparency through detailed experimental provenance; and (4) enable reproducible, comparative analysis of method robustness. By making these tools openly available within \skc, we address the critical gap between theoretical rank reversal criteria and practical reliability assessment, facilitating broader adoption of systematic robustness analysis in \mcda applications across diverse domains. Philosophically, this work adopts a pragmatic perspective: by providing comprehensive tools to measure, quantify, and compare rank reversal behavior across methods, we enable researchers and practitioners to make informed decisions about method selection based on empirical robustness analysis.

The remainder of this paper is organized as follows. Section~\ref{section:from-theory} examines why a validated computational implementation of Wang and Triantaphyllou's criteria has remained absent from the literature for 17 years, framing the specific algorithmic challenges addressed in this work. Section~\ref{section:rankcmp} establishes the conceptual foundations by introducing the \texttt{RankResult} and \texttt{RanksComparator} data structures that underpin our comparative analysis framework. Sections~\ref{section:rrt1},~\ref{section:rrt2}, and~\ref{sec:rrt3} present detailed implementations of the three rank reversal tests, including algorithmic specifications and methodological considerations. Section~\ref{section:case-studies} illustrates the practical applicability of these tests through two case studies: an in-depth application to a single dataset, and a large-scale audit of published \mcdm examples. Finally, Section~\ref{section:summary} synthesizes our contributions and discusses implications for \mcda reliability assessment and method comparison.

\section{From Theory to Implementation: Algorithmic Challenges}
\label{section:from-theory}

While \citet{wang2009rank} expanded this theory to multiple
methods, and alternatives such as \textsc{comet}, \textsc{spotis}, and
\textsc{rafsi} proposed reversal-resistant approaches
\citep{spotisCometRafsiRRR2021}, none provided computational tools to
\textit{detect} the phenomenon. A separate, earlier detection criterion, based on
introducing a clone of a non-optimal alternative and checking whether
the best alternative remains unchanged, was proposed by \citet{belton1983short};
we have also implemented this test in \skc
(\texttt{RankClonesChecker}) as a complementary, lighter-weight check
alongside the three Wang and Triantaphyllou criteria that are the focus
of this paper, and we leave a detailed treatment of it for future work.

Our informal survey of over 100 GitHub repositories, 15+ PyPI packages, 10+ CRAN packages, 20+ MATLAB File Exchange submissions, and commercial software (searched using combinations of the terms ``rank reversal'', ``Wang Triantaphyllou'', and ``MCDA robustness testing'') found no existing tool that fully implements Wang's three tests.
This gap persists despite explicit recognition in recent literature:
\citet{aires2018rank} documented that studies on rank reversal remain scarce,
noting that practical application has been
limited by the absence of comprehensive computational implementations. The
17-year duration of this gap is notable when compared to other methodological
advances, and reflects the substantial technical
complexity involved in translating abstract theoretical definitions into
concrete algorithms, a challenge requiring expertise in both decision theory
and software engineering.

The complexity of this transformation becomes evident when comparing Wang's
conceptual guidelines with the idea of translating them into a concrete
implementation. Test 1, originally defined as ``degrading suboptimal
alternatives'', requires developing a degradation algorithm with statistical
control limits. Aspects not contemplated in the original work, such as tie
handling, alternatives eliminated by preprocessing filters, and integration
with complex \mcda pipelines, demand additional solutions: a hierarchical
deterministic tie-breaking system, a graceful degradation strategy with ranking
assignment, and full support for preprocessing stages. Tests 2 and 3, originally formulated in terms of pairwise comparisons and
consistent recomposition respectively, can be implemented through the
construction of dominance graphs with cycle detection, followed by exact
graph condensation and transitive reduction to derive rankings without
discarding any pairwise evidence. This transformation of abstract concepts
into robust computational procedures represents a genuine algorithmic
contribution that goes beyond mere ``coding''.

\section{Our Technical Foundations: Ranks and Ranks-Comparators}
\label{section:rankcmp}

Within the \skc framework, two fundamental data structures support
comparative ranking analysis: \texttt{RankResult} and
\texttt{RanksComparator}. These structures form the computational foundation
upon which the rank reversal detection criteria of Wang and Triantaphyllou
are practically realized.

The \texttt{RankResult} class encapsulates the output of a multi-criteria
decision method (\mcdm) that produces an ordered ranking of alternatives. It
handles ordinal classifications of alternatives (where lower values indicate
higher preference) and incorporates metadata about the generating method,
support for intermediate calculations, ranking manipulation, and tie handling.

\texttt{RanksComparator}, in turn, emerged from the need to compare multiple
rankings generated by different methods or configurations applied to the same
decision problem \citep{cabral2025beyond}. It implements an iterable interface
and provides statistical calculations (correlations, covariances, coefficients
of determination, and distances between rankings), specialized visualizations,
and quantitative measures of method stability under controlled perturbations.
Both classes also incorporate a flexible metadata container that enhances the
transparency and traceability of \mcdm processes. In the analyses that follow,
this container is used to systematically record mutation information: iteration
numbers, modified alternatives, applied noise vectors, and missing alternative
tracking. This supports reproducibility and the scientific rigor of the
evaluation pipeline.

Together, these structures provide the analytical framework for comparing and
critically evaluating different classifications of alternatives, and serve as
the conceptual foundation for the rank reversal tests described in the
following sections.

\section{Rank Reversal Test 1: A Systematic Approach to Alternative Replacement}
\label{section:rrt1}

The first criterion in Wang and Triantaphyllou's framework, the Rank Reversal
Test 1 (\textsc{rrt}1), addresses a fundamental question about \mcda method reliability:
does the optimal alternative remain stable when suboptimal alternatives are
systematically degraded? \citep{wang2008ranking}. Our implementation transforms
this theoretical criterion into a concrete algorithmic framework through
controlled mutation experiments. Rather than arbitrary modifications, we employ
a systematic deterioration strategy that preserves the logical structure of the
problem while evaluating method stability. Algorithm~\ref{alg:rrt1} presents
the core logic of our approach.

\begin{algorithm}
\caption{Implementation of Wang-Triantaphyllou's \textsc{rrt}1 criterion: Systematic degradation of suboptimal alternatives with multiple repetitions for comprehensive rank reversal testing}
\label{alg:rrt1}
\input{algorithm/rrt1}
\end{algorithm}

This algorithm operates through three conceptual phases that collectively
address the fundamental challenge of systematic rank reversal detection.

\textbf{Phase 1: Baseline establishment}. Begin by evaluating
the original decision matrix using the target \mcda method, establishing a
reference ranking. This baseline serves as the stability anchor against which
all subsequent mutations are compared. The preservation of the optimal
alternative from this baseline constitutes the core stability criterion.

\textbf{Phase 2: Systematic mutation experimentation}. The algorithm enters
its core experimental phase, implementing a nested iteration structure that
ensures comprehensive coverage of the decision space. The outer loop controls
experimental repetition to enable statistical analysis of stability, while the
inner loop systematically targets each suboptimal alternative for degradation.
This design ensures that every suboptimal alternative undergoes controlled
mutation testing across multiple experimental trials. Throughout this process,
the algorithm maintains detailed provenance information about each mutation,
enabling full traceability of experimental conditions.

\textbf{Phase 3: Results integration}. The integration of all rankings into a
\texttt{RanksComparator} object transforms the experimental results into an
analytically rich data structure compatible with the broader comparative
analysis framework.

Translating these phases in practice is not trivial. Generating controlled
yet meaningful degradations, handling alternatives eliminated by preprocessing
pipelines, and ensuring full experimental traceability constitute the main
technical challenges of this implementation.

\subsubsection{Degradation Strategy and Noise Generation}

The algorithm operates on a straightforward principle: systematically worsen
each suboptimal alternative while maintaining ordinal consistency. Given an
original ranking

\begin{equation}
A_1 \succ A_2 \succ A_3 \succ A_4 \succ A_5
\end{equation}

for each $A_i$ where $i > 1$, we generate a degraded alternative $A'_i$ such
that:

\begin{equation}
A_{i-1} \succ A'_i \succ A_{i+1}
\end{equation}

This ensures that mutations are both meaningful, representing realistic
degradations, and bounded, without violating the existing preference structure.
Since \mcda problems are inherently multidimensional, achieving this ordinal
constraint requires a noise generation procedure that operates across all
criteria simultaneously:

\begin{enumerate}
    \item \textbf{Inter-alternative difference estimation}: For each suboptimal
    alternative $A_k$, compute the absolute difference with the next-worse
    alternative $A_{k+1}$ across all criteria $j$:
    \begin{equation}
        \delta_{k,j} = |A_{k,j} - A_{k+1,j}|
    \end{equation}

    \item \textbf{Bounded noise sampling}: For each criterion $j$, sample
    noise from a uniform distribution bounded by the estimated difference:
    \begin{equation}
        \epsilon_{k,j} \sim \mathcal{U}[0, \delta_{k,j}]
    \end{equation}

    In practice, $\delta_{k,j}$ is further clipped to
    $\min(\delta_{k,j}, |A_{k,j}|)$ before sampling, so that
    $A_{k,j} - \epsilon_{k,j}$ can approach zero but never cross it. Without
    this safeguard, criteria with highly heterogeneous scales across
    alternatives could sample noise larger than the value being perturbed,
    flipping the sign of that value and potentially changing the semantic meaning of the criterion.

    \item \textbf{Last alternative handling}: The worst-ranked alternative
    $A_n$ has no natural lower bound, so its noise limit is derived from the
    median of all pairwise differences:
    \begin{equation}
        \epsilon_{n,j} \sim \mathcal{U}\left[0, \text{median}\left(\{\delta_{k,j}\}_{k=1}^{n-1}\right)\right]
    \end{equation}
    The median is preferred over the mean for its robustness against outlier
    gaps between adjacent alternatives.

    This median-based bound is a pragmatic, simple choice rather than one
    derived analytically from $A_{n,j}$'s own value: it is instead computed
    as an aggregate statistic over the \emph{other} alternatives, so the
    sampled range is not directly tied to the magnitude being degraded.
    The sign-flip clip described above still applies and prevents this
    from causing a zero-crossing in practice. Still, anchoring the bound
    to $A_{n,j}$ directly (e.g., as a fraction of $|A_{n,j}|$) remains an
    open refinement for future work.

    \item \textbf{Directional adjustment}: Apply $\epsilon_{k,j}$ negatively
    to maximization criteria and positively to minimization criteria.
\end{enumerate}

This strategy ensures that every mutated alternative $A'_k$ satisfies
$A_{k-1} \succ A'_k \succ A_{k+1}$, preserving ordinal relationships while
introducing controlled degradation.

\subsubsection{Pipeline Compatibility and Graceful Degradation}

\skc frequently employs complex preprocessing pipelines that combine multiple
transformation steps before the final evaluation. Following the composition
paradigm established in Scikit-Learn \citep{pedregosa2011scikit}, the project
implements \texttt{SKCPipeline} objects that chain transformers and aggregation
functions into a unified decision-maker.

These pipelines may incorporate two theoretically grounded filtering mechanisms
that are particularly relevant for \textsc{rrt}1:

\begin{enumerate}
    \item \textbf{Satisficing analysis}: Alternatives that do not meet minimum
    performance thresholds on critical criteria are eliminated early in the
    process, following Simon's satisficing theory.
    \item \textbf{Dominance analysis}: The Pareto optimality principle dictates
    that dominated alternatives must be excluded from further consideration,
    as they can never be optimal under any reasonable
    preference structure.
\end{enumerate}

A typical \skc pipeline is shown in Code~\ref{code:pipeline}.

\begin{figure}[t]
\hrule height 1pt width \linewidth
\begin{lstlisting}[language=Python, caption={Multi-criteria decision pipeline
construction using \skc. The pipeline includes: (1) satisficing filter with
$criteria \leq 1000$, (2) dominance-based filtering, (3) weight normalization
via sum scaling, (4) matrix normalization via vector scaling, and (5) final
evaluation with \textsc{topsis}.}, label=code:pipeline]
from skcriteria.pipeline import mkpipe
pipeline = mkpipe(
   InvertMinimize(), # convert minimization criteria to maximization
   FilterGT({'criteria': 1000}), # remove alternatives below threshold
   FilterNonDominated(),   # keep only non-dominated alternatives
   SumScaler(target="weights"), # normalize weights to sum 1
   VectorScaler(target="matrix"), # normalize matrix by unit vectors
   TOPSIS() # rank by distance to ideal solution
)
\end{lstlisting}
\end{figure}

When a pipeline eliminates alternatives during preprocessing, a methodological
challenge arises: \textsc{rrt}1 must evaluate ranking stability when alternatives are
mutated, but some may have been removed by theoretically valid filtering
processes. This is further compounded by a design constraint of
\texttt{RanksComparator}, which requires all rankings to contain exactly the
same set of alternatives \citep{cabral2025beyond}. Our implementation resolves both tensions through
\textit{graceful degradation}, a concept from systems engineering whereby a
system maintains partial functionality when some components fail.
Algorithm~\ref{alg:missing-alternatives} details
its implementation.

\begin{algorithm}
\caption{Graceful degradation algorithm for missing alternatives: detection of
pipeline-eliminated alternatives and assignment of the worst possible rank
(max\_rank + 1) to maintain ranking completeness}
\label{alg:missing-alternatives}
\input{algorithm/graceful_degradation}
\end{algorithm}

This approach recognizes that alternatives eliminated through satisficing or
dominance analysis would naturally rank last in any comprehensive evaluation,
justifying their assignment to the worst ranking positions. Since our
representation supports ties, multiple eliminated alternatives can share the
same worst rank, preserving the mathematical consistency of the ranking
structure. Nevertheless, to allow flexibility in pipeline design, our
implementation also supports a stricter mode via the \texttt{allow\_missing=False}
parameter, which raises an error if the pipeline eliminates any alternative
during evaluation. This option is useful when the analyst requires that all
alternatives remain present throughout the entire decision process, ensuring
that rank reversal testing operates on a complete and unmodified alternative
set.

\subsubsection{Output and Interpretation}

The \textsc{rrt}1 returns a \texttt{RanksComparator} object consolidating the original
ranking alongside all mutation rankings, leaving the stability assessment to
the analyst rather than producing a binary pass/fail verdict; the full
analytical infrastructure of \texttt{RanksComparator} (dataframe export,
distance metrics, and flow/box plots) remains available for this purpose.
The \texttt{extra\_} attribute of the returned object stores the following
diagnostic information:

\begin{itemize}

    \item \textbf{Optimal Preservation Rate (OPR)}: The proportion of mutations in
    which the optimal alternative $a^*$ retains its top position. Let
    $\mathcal{M}^* = \{m \in \mathcal{M} : r_m(a^*) = 1\}$ be the set of
    mutations where $a^*$ remains optimal. Then:
    $$
    \text{OPR} = \frac{|\mathcal{M}^*|}{|\mathcal{M}|}
    $$
    A value of 1 indicates perfect stability; 0 indicates that $a^*$ is
    displaced in every mutation.

    \item \textbf{Rank Displacement (RD)}: The average magnitude of displacement
    of $a^*$ across all mutations:
    $$
    \text{RD} = \frac{1}{|\mathcal{M}|} \sum_{m \in \mathcal{M}} (r_m(a^*) - 1)
    $$
    where $r_m(a^*)$ is the rank assigned to $a^*$ in mutation $m$. A value
    of 0 confirms that $a^*$ consistently holds the top position, while larger
    values indicate increasingly severe instability.
\end{itemize}

Additionally, each individual ranking in the \texttt{RanksComparator} stores
detailed mutation provenance in its own \texttt{extra\_.rank\_inv\_check}
attribute, including the degraded alternative, iteration number, exact noise
vector applied to each criterion, and any alternatives filtered by the pipeline.

% ====================================================

\section{Rank Reversal Test 2: Transitivity Analysis through Pairwise
Decomposition}
\label{section:rrt2}

\begin{figure}[h]
\centering
\includegraphics[width=\textwidth]{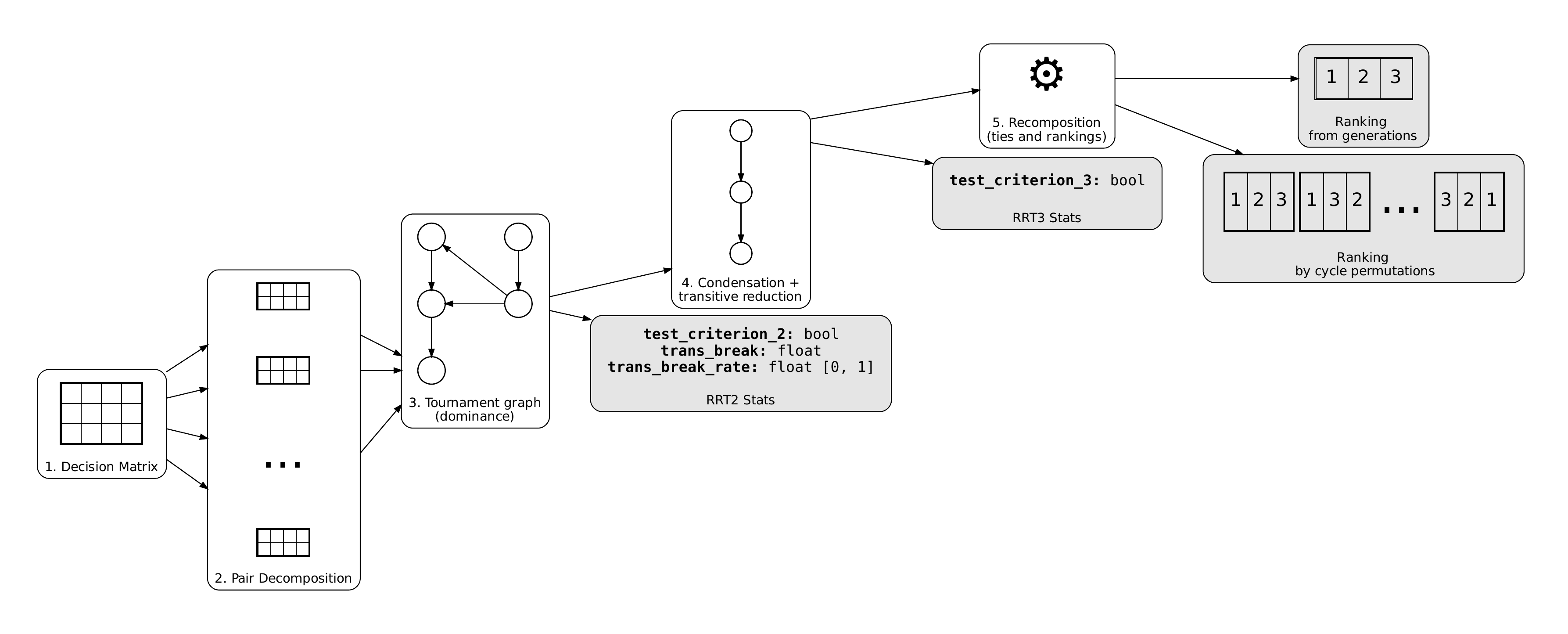}
\caption{Conceptual overview of the \textsc{rrt}2/\textsc{rrt}3 pipeline. (1) The decision matrix is
(2) decomposed into all $n(n-1)/2$ alternative pairs, evaluated in parallel,
and (3) aggregated into a dominance graph (an $n$-tournament), from which
\textsc{rrt}2's diagnostics (RTT2 stats) are computed. (4) The graph is condensed and
transitively reduced, collapsing dominance cycles into explicit ties, and
(5) recomposed into either a tied ranking (\texttt{generations}) or all
consistent strict orderings (\texttt{cycle\_permutations}); \textsc{rrt}3's
\texttt{test\_criterion\_3} reports whether this recomposition matches the
original ranking. Gray-shaded boxes denote the pipeline's outputs.}
\label{fig:rank_pipeline}
\end{figure}

{Unlike \textsc{rrt}1, which reduces to a single mutate-and-reevaluate loop, \textsc{rrt}2
and \textsc{rrt}3 involve a multi-stage pipeline spanning pairwise decomposition,
dominance graph construction, cycle detection, condensation, and ranking
recomposition. Given this added conceptual complexity, we first} present a
conceptual, high-level summary of the complete \textsc{rrt}2/3 pipeline in
Figure~\ref{fig:rank_pipeline} before introducing the underlying
mathematical and algorithmic details: the remainder of this section develops
stages (1)-(3) (pairwise decomposition, dominance graph construction, and
transitivity verification), while Section~\ref{sec:rrt3} develops stages
(4)-(5) (graph condensation and ranking recomposition).

The second criterion in Wang and Triantaphyllou's framework (\textsc{rrt}2) addresses
the logical consistency of rankings through transitivity: if pairwise
comparisons indicate that $A \succ B$ and $B \succ C$, then $A \succ C$ must
hold \citep{wang2008ranking}. A violation of this property forms a preference
cycle, contradicting the fundamental axioms of rational decision-making.
While the theoretical requirement is clear, its implementation raises
non-trivial computational challenges: the original formulation does not specify
how pairwise comparisons should be reconstructed or aggregated into a global
preference order. Our approach addresses this by constructing a dominance graph
from all pairwise comparisons and analyzing the consistency of alternative
orderings through topological analysis methods. Algorithm~\ref{alg:rrt2}
presents the core logic of our approach.

\begin{algorithm}[t]
\caption{Proposed implementation of \textsc{rrt}2: Construction of the dominance graph from pairwise comparisons and detection of simple cycles (closed paths with no repeated nodes) representing transitivity violations}
\label{alg:rrt2}
\input{algorithm/rrt2}
\end{algorithm}

The algorithm operates in three phases that ensure transitivity is preserved:

\textbf{Phase 1: Baseline Establishment.}
The algorithm establishes a reference ranking by evaluating the original decision matrix using the target \mcdm method or pipeline. This ranking serves as a transitivity anchor against which all pairwise decomposition results are compared.
This reference is used as part of Test 3, which will be explained in Section~\ref{sec:rrt3}.

\textbf{Phase 2: Pairwise Decomposition and Graph Construction.}
The algorithm decomposes the original decision problem into all possible pairwise comparisons.
Thus, for $n$ alternatives, we generate ${n(n-1)}/{2}$ subproblems, each with a reduced decision matrix $D_{ij}$, restricted to only two alternatives $(A_i, A_j)$ while preserving the criteria structure.
Each subproblem is evaluated independently and in parallel using the chosen \mcdm method/pipeline. The results are aggregated into a directed graph $G = (V, E)$, where nodes $V$ represent alternatives and an edge $(A_i, A_j) \in E$ indicates that $A_i$ is preferred over $A_j$.

This representation enables structural analysis of preference relations using graph theory tools. In particular, the presence of directed cycles in this graph constitutes a violation of the transitivity principle.

\textbf{Phase 3: Transitivity Verification.}
\label{phase:3}
Transitivity violations manifest as directed cycles in the preference graph, and to detect them, we identify all directed cycles of length 3, which correspond to basic transitivity violations. We then define the transitivity violation rate as:

$$
\text{Transitivity Violation Rate} = \frac{\text{Number of 3-cycles detected}}{\text{Maximum possible 3-cycles}}
$$

Since the resulting graph is an $n$-tournament (a directed graph with exactly one edge between each pair of vertices oriented in one of the two possible directions), according to the Corollary of Theorem 4 by Moon \citep{moon1968topics}, the denominator is given by:

$$
\text{Maximum possible 3-cycles} =
\begin{cases}
{n(n^2-4)}/{24} & \text{if } n \text{ is even} \\
{n(n^2-1)}/{24} & \text{if } n \text{ is odd}
\end{cases}
$$

This normalization facilitates comparison across problems of different sizes.

Finally, the transitivity validation is considered satisfied if and only if the violation rate is zero.

\subsubsection{Output and Interpretation}

The \textsc{rrt}2 returns a \texttt{RanksComparator} object consolidating the original
ranking alongside all reconstructed rankings. The \texttt{extra\_} attribute
of the returned \texttt{RanksComparator} stores: the Boolean \textbf{test
criterion 2 result} (whether the Transitivity Violation Rate equals zero);
the \textbf{pairwise dominance graph}, stored as a NetworkX \texttt{DiGraph}
for further structural analysis; the list of \textbf{transitivity violations}
(the 3-cycles $(A, B, C)$ where violations occur, formatted as
human-readable strings such as $A>B>C>A$), together with the normalized
\textbf{Transitivity Violation Rate} defined in Phase 3; and the complete
list of \textbf{pairwise comparisons}, enabling detailed inspection of
individual preference relationships.

\textsc{rrt}2 and \textsc{rrt}3 are exposed together: a single
call to \texttt{evaluate} performs the pairwise
decomposition, builds the dominance graph, and evaluates transitivity (\textsc{rrt}2)
as described above; it then reuses that same graph to perform condensation
and ranking recomposition, evaluating recomposition consistency (\textsc{rrt}3,
Section~\ref{sec:rrt3}). Both sets of diagnostics are consolidated into the
same \texttt{RanksComparator} object. We keep the exposition split across two
sections for clarity, but every item listed above and in
Section~\ref{sec:rrt3_output} is available simultaneously in the same object.

\section{Rank Reversal Test 3: Recomposition Consistency}
\label{sec:rrt3}

The third criterion in Wang and Triantaphyllou's framework (\textsc{rrt}3) evaluates
whether the original ranking retains its structure when reconstructed from
the pairwise dominance graph built in \textsc{rrt}2 \citep{wang2008ranking}. We refer
to this property as \textit{recomposition consistency}: the test is satisfied
if and only if the reconstructed ranking is identical to the original, and
fails if any alternative changes its position after reconstruction.

A stricter requirement underlies this criterion: the recomposed ranking must
be unique. \textsc{rrt}3 is satisfied only when the pairwise sub-problems yield a
single valid topological ordering that coincides with the original ranking.
If multiple valid orderings exist, the recomposition is ambiguous and \textsc{rrt}3
fails, regardless of whether any of those orderings happens to match the
original.

When the dominance graph obtained from \textsc{rrt}2 is acyclic, it defines a strict
total order \citep{moon1968topics} with a unique topological ordering,
making recomposition unambiguous. However, when \textsc{rrt}2 detects transitivity
violations in the form of directed cycles, a unique total order cannot be
derived. In this case, \textsc{rrt}3 is evaluated over a set of candidate rankings
obtained by condensing the dominance graph's cycles into explicit ties and
applying the ranking recomposition strategies described in the following
subsections. This approach allows \textsc{rrt}3 to remain informative even when transitivity is not fully satisfied, providing a richer picture of the decision method's consistency under different recomposition scenarios.

\subsection{DAG Conversion via Graph Condensation}
\label{sec:condensation}

To address this challenge, our current implementation abandons cycle-breaking
heuristics altogether in favor of an exact, deterministic graph transformation:
\textit{condensation}. Given the dominance graph $G = (V, E)$ built in \textsc{rrt}2, we
apply condensation to obtain the graph $G^{\ast}$, which is always acyclic
regardless of how many or how large the cycles in $G$ were
\citep{tarjan1972depth}. We then apply \textit{transitive reduction} to
$G^{\ast}$, which drops every edge that is already implied by a longer path
between the same two supernodes, leaving only the minimal set of edges
required to preserve the reachability relation. Both steps are provided
directly by \texttt{NetworkX} \citep{hagberg2007exploring}.

This design discards no information: every alternative that participated in a
cycle is preserved, together with all of the alternatives it was cyclically
tied to, inside its supernode. No edge is removed and no arbitrary decision is
required to choose which preference relation to sacrifice. The procedure is
always exact, runs in $O(|V| + |E|)$ time, where $|V|$ and $|E|$ represent the
number of nodes/alternatives and edges/dominance relations of the graph,
respectively, and is fully deterministic: the same
dominance graph always yields the same condensation. Rather than
\textit{forcing} a strict order where the pairwise evidence does not support
one, this approach \textit{reports} the ambiguity as an explicit tie, leaving
its resolution, if one is desired at all, to the ranking recomposition
strategies described in Section~\ref{sec:recomposition}.

\begin{figure}[t]
\centering
\includegraphics[width=\columnwidth]{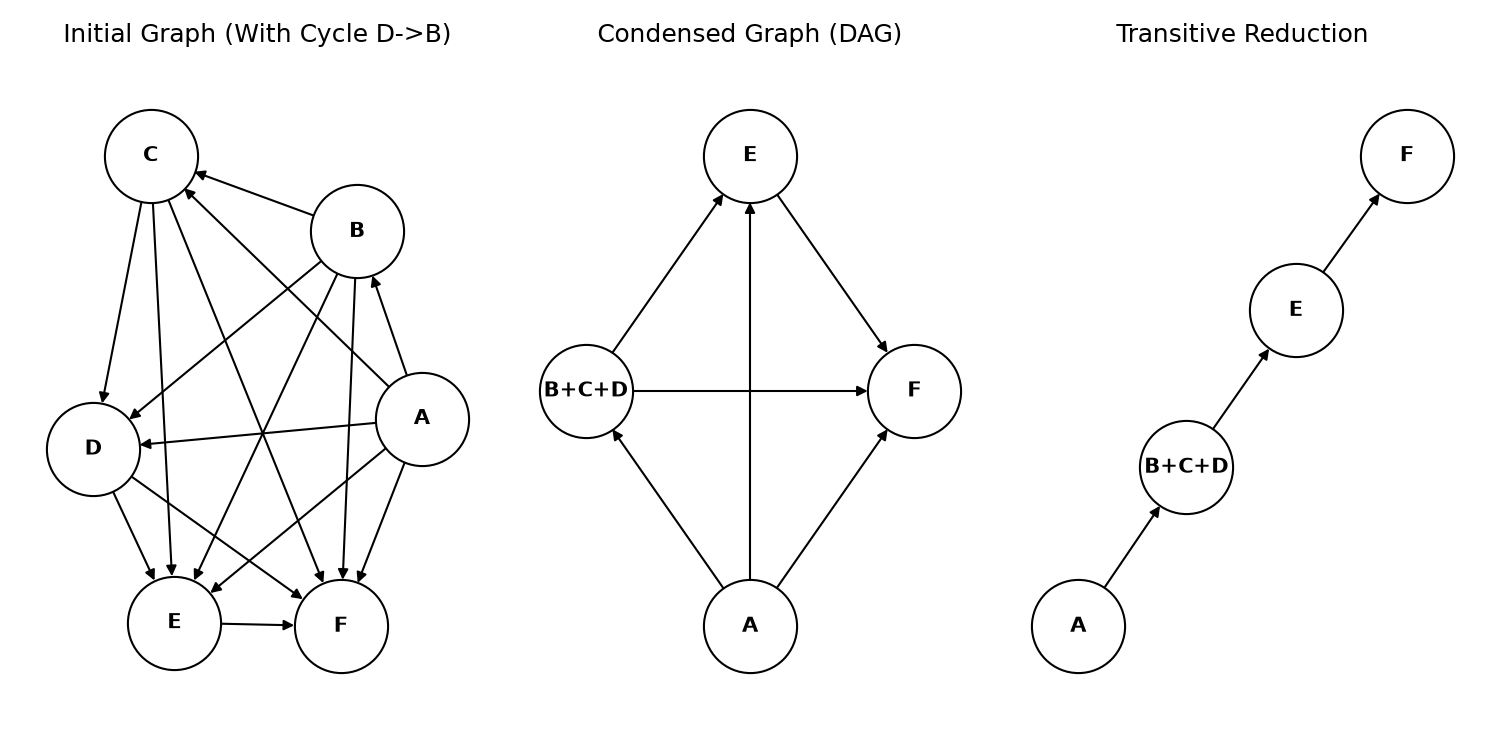}
\caption{{Condensation and transitive reduction of a dominance graph containing a cycle $B \succ C \succ D \succ B$. \textbf{Left:} initial dominance graph, hand-constructed to contain this cycle. \textbf{Center:} the condensed graph $G^{\ast}$ (output of \texttt{nx.condensation}), where $B$, $C$, $D$ collapse into a supernode representing their tie. \textbf{Right:} the transitively reduced graph (output of \texttt{nx.transitive\_reduction}), retaining only the minimal edges needed to preserve reachability.}}
\label{fig:dag-vs-cyclic}
\end{figure}

Figure~\ref{fig:dag-vs-cyclic} illustrates this procedure on a worked example.
{The initial dominance graph (left) is a hand-constructed input, built
deliberately to contain a cycle $B \succ C \succ D \succ B$ for illustration
purposes; the condensed graph $G^{\ast}$ (center) and its transitively
reduced counterpart (right), by contrast, are not schematics but the direct
output of executing our implementation (\texttt{nx.condensation} followed by
\texttt{nx.transitive\_reduction}) on that input graph.} From this cycle, no
strict ranking can be derived among $B$, $C$, and $D$; in $G^{\ast}$, $B$,
$C$, and $D$ are collapsed into a single supernode representing their mutual
tie, and the transitively reduced graph retains only the minimal set of
edges needed to preserve the reachability relation. This procedure is applied
uniformly regardless of whether the dominance graph contains cycles: when $G$
is already acyclic, every \textsc{scc} is a singleton, and the condensation
(followed by transitive reduction) simply reproduces $G$ without altering its
structure, so no special-casing is required between the cyclic and acyclic
scenarios.

\subsection{Ranking Recomposition Strategies}
\label{sec:recomposition}

Once the dominance graph has been condensed and transitively reduced,
rankings are derived through an analysis of the resulting DAG. Our
implementation provides two complementary strategies, each offering
different trade-offs between compactness and exhaustiveness, such that every
alternative belonging to a collapsed supernode is correctly accounted for.

The \textbf{generations} strategy produces a single ranking based on
topological generations, i.e., the layers of the condensed DAG where
supernodes are mutually incomparable. All alternatives that belong to the
same supernode, whether a singleton or a full dominance cycle collapsed
during condensation, receive the same rank, resulting in a ranking with ties
that directly and transparently reflect the ambiguity detected in \textsc{rrt}2,
rather than an artifact of an arbitrary cycle-breaking choice. This strategy
is computationally efficient ($O(|V| + |E|)$) and provides a canonical
representation that explicitly captures preference indifference between
incomparable alternatives. Algorithm~\ref{alg:generations} details its
implementation.

\begin{algorithm}[t]
\caption{Ranking from Topological Generations of the Condensed DAG}
\label{alg:generations}
\begin{algorithmic}[1]
\Require Alternatives array $A$, condensed reduced DAG $G'$, members map $M$
\Ensure Ranking array $R$ (1-indexed, with ties for cyclically-tied alternatives)
\State $rank \gets 1$
\For{each generation $gen$ in topological\_generations$(G')$}
    \For{each supernode $s \in gen$}
        \For{each alternative $a \in M[s]$}
            \State $\text{alt\_to\_rank}[a] \gets rank$
        \EndFor
    \EndFor
    \State $rank \gets rank + 1$
\EndFor
\State $R \gets [\text{alt\_to\_rank}[a]$ for $a \in A]$
\State \Return $R$
\end{algorithmic}
\end{algorithm}

The \textbf{cycle\_permutations} strategy takes the opposite approach:
instead of hiding the ambiguity behind a tie, it enumerates every internally
plausible strict ordering of each cycle's members as a separate candidate
ranking.
{This relies on the fact that the dominance graph is always an $n$-tournament by construction, and the condensation of a
tournament is always a \textit{strict total order} of its supernodes, i.e.,
no two supernodes are ever incomparable}.
Consequently, every topological
generation of the condensed DAG contains exactly one supernode, which means
the only combinatorial freedom left comes from permuting the alternatives
\textit{within} each supernode. This lets the algorithm enumerate the
Cartesian product of the internal permutations of every supernode directly.
The total number of rankings produced is therefore exactly
$\prod_{s} |M[s]|!$, the product, over every supernode $s$, of the factorial
of its size, rather than a bound that depends on the interleaving of
independent branches of a general DAG. A cycle of $k$ tied alternatives
therefore contributes exactly $k!$ candidate rankings, while singleton
supernodes contribute a single, fixed position. Note that, since the
factorial function is convex, this product is maximized when the size is
concentrated in the smallest possible number of supernodes; the absolute
worst case occurs when all $|V|$ alternatives collapse into a single cycle
involving every one of them, producing $|V|!$ rankings. Since this count can
grow quickly for large cycles, the implementation exposes a
\texttt{max\_rankings} parameter to cap the number of rankings generated
(default: 50 when invoked from \texttt{RankTransitivityChecker}).
Algorithm~\ref{alg:cycle-permutations} details its implementation.

\begin{algorithm}[t]
\caption{Generate Rankings from Cycle Permutations}
\label{alg:cycle-permutations}
\begin{algorithmic}[1]
\Require
Alternatives array $A$, condensed reduced DAG $G'$, members map $M$,
    max\_rankings $N$
\Ensure Generator yielding ranking arrays (strict total orders, no ties)
\State $\text{all\_permutations} \gets []$
\For{each generation $gen$ in topological\_generations$(G')$}
    \State $s \gets$ the single supernode in $gen$
    \State $\text{all\_permutations.append}(\text{permutations}(M[s]))$
\EndFor
\State $count \gets 0$
\For{each $\text{combination}$ in $\text{product}(\text{all\_permutations})$}
    \If{$N \neq \text{None}$ and $count \geq N$}
        \State \textbf{break}
    \EndIf
    \State $\text{full\_order} \gets \text{concatenate}(\text{combination})$
    \State $\text{alt\_to\_rank} \gets \{a: i+1$ for $i, a$ in
        enumerate(full\_order)$\}$
    \State $R \gets [\text{alt\_to\_rank}[a]$ for $a \in A]$
    \State \textbf{yield} $R$
    \State $count \gets count + 1$
\EndFor
\end{algorithmic}
\end{algorithm}

The complete \textsc{rrt}3 procedure, integrating condensation-based DAG conversion and
ranking recomposition, is presented in Algorithm~\ref{alg:rrt3}.

\begin{algorithm}[H]
\caption{\textsc{rrt}3: Recomposition Consistency Test}
\label{alg:rrt3}
\begin{algorithmic}[1]
\Require Test criterion 2 result, original ranking $R_{orig}$, dominance
    graph $G$, ranking strategy $\in \{\text{generations}, \text{cycle\_permutations}\}$,
    max\_rankings $N$
\Ensure Test criterion 3 result, reconstructed rankings, condensed reduced DAG
\State $(G', M) \gets \text{as\_condensed\_reduced\_dag}(G)$
\State $reconstructed \gets []$
\If{strategy $=$ generations}
    \State $R_{gen} \gets \text{ranking\_from\_generations}(A, G', M)$
    \State $reconstructed.\text{append}(R_{gen})$
\ElsIf{strategy $=$ cycle\_permutations}
    \For{each $R$ in generate\_rankings\_with\_cycle\_permutations$(A, G', M, N)$}
        \State $reconstructed.\text{append}(R)$
    \EndFor
\EndIf
\State $test\_criterion\_3 \gets test\_criterion\_2$ \textbf{and}
    $(R_{orig} = reconstructed[0])$
\State \Return $(test\_criterion\_3, reconstructed, G')$
\end{algorithmic}
\end{algorithm}

\subsection{Output and Interpretation}
\label{sec:rrt3_output}

As noted in Section~\ref{section:rrt2}, \textsc{rrt}2 and \textsc{rrt}3 are evaluated together
and consolidated into the same \texttt{RanksComparator} object. The \textsc{rrt}3
diagnostics extend that object with the original ranking alongside all
reconstructed rankings. The test criterion is satisfied if and only if \textsc{rrt}2
passed and the original ranking coincides exactly with the first
reconstructed ranking. Even when \textsc{rrt}3 fails, the reconstructed rankings
provide valuable diagnostic information, allowing the analyst to assess the
degree of recomposition ambiguity and the severity of ranking inconsistencies
through the full analytical infrastructure of \texttt{RanksComparator}.

The \texttt{extra\_} attribute of the returned \texttt{RanksComparator}
additionally stores: the Boolean \textbf{test criterion 3 result}; the
\textbf{condensed reduced DAG} obtained by collapsing every dominance cycle
detected in \textsc{rrt}2 into a supernode and applying transitive reduction,
stored as a NetworkX \texttt{DiGraph} whose nodes carry the alternatives
they represent; the full collection of \textbf{reconstructed rankings}
derived from the condensed DAG under the selected \texttt{ranking\_strategy}
(\texttt{generations} or \texttt{cycle\_permutations}); and the
\textbf{maximum possible rankings (\texttt{mpr})} derivable from the
condensed DAG (Section~\ref{sec:recomposition}), reported regardless of the
selected strategy so the analyst can assess whether \texttt{max\_rankings}
truncated the \texttt{cycle\_permutations} enumeration.

Additionally, each individual reconstructed ranking stores, in its own
\texttt{extra\_\\.transitivity\_check} attribute, the alternatives that were
missing from the decision maker's output (if
\texttt{allow\_missing\_alternatives=True}) and a \texttt{recomposi\\tion}
identifier: the literal string \texttt{"generations"} for the generations
strategy, or an integer index identifying which cycle-permutation combination
produced that particular ranking. Because condensation neither removes edges
nor makes arbitrary choices, this metadata is sufficient to fully retrace,
for any reconstructed ranking, which dominance cycles were tied and which
internal ordering of each cycle was used to produce it.

\subsection{Computational Complexity}
\label{sec:complejidad}

The dominant cost is the $O(n^2)$ pairwise evaluations, which can be
mitigated through parallel execution (\texttt{preferred\_parallel\_backend}
and \texttt{n\_jobs} parameters); since each pairwise comparison is
independent of the others, this workload is embarrassingly parallel.
Condensation and transitive reduction are exact and computationally
inexpensive even for large problems, since their cost depends only on the
size of the dominance graph and not on the number or size of the detected
cycles. For problems with more than 50 alternatives, we recommend the
generations strategy for recomposition, since \texttt{cycle\_permutations}
only becomes costly when large dominance cycles are present (in the absence
of cycles, both strategies are equivalent in cost, as each
$|M[s]|! = 1$).

\section{Case Studies: Illustrative Applications}
\label{section:case-studies}

Having presented \textsc{rrt}1, \textsc{rrt}2, and \textsc{rrt}3 as algorithmic tools, we now
illustrate their practical use through two complementary case studies.
The first is a focused, single-dataset application that exercises both
families of tests (\textsc{rrt}1 and \textsc{rrt}2/\textsc{rrt}3) side by side under two
experimental configurations, showing in detail how the diagnostics
introduced in the previous sections are computed and interpreted on a
concrete decision problem. The second broadens the scope considerably,
applying the same tests to a diverse sample of 27 pipeline/dataset
combinations drawn directly from the published \mcdm literature, in
order to gauge how often rank reversal issues actually arise ``in the
wild'' rather than in adversarially constructed examples.

Together, these two cases are intended to demonstrate applicability
rather than to introduce new methodology: the first shows \textit{how} to
read and combine the \textsc{rrt}1-\textsc{rrt}3 diagnostics on a single, realistic
problem, while the second shows \textit{why} this matters empirically,
by quantifying how frequently each criterion is violated across a
representative slice of the field. Both analyses were carried out with
the implementation described in the previous sections, and reuse the
same \texttt{RankInvariantChecker} and \texttt{RankTransitivityChecker}
interfaces without any case-specific modification.

The complete experiments, including the full analysis notebooks, code,
and additional plots not reproduced here for brevity, will be made
publicly available at \url{https://github.com/quatrope/skcriteria_rankrev_paper}.

\subsection{Case 1: \textsc{topsis} Robustness on a Cryptocurrency Evaluation Dataset}
\label{sec:case1-crypto}

The first case study uses the cryptocurrency evaluation dataset of
\citet{vanheerden2021crypto} with a \textsc{topsis} pipeline similar
to the one used in that work
(\texttt{NegateMinimize} $\to$ \texttt{EntropyWeighter} $\to$
\texttt{VectorScaler} $\to$ \texttt{TOPSIS}), but replacing their
approach to converting minimization criteria to maximization with
\texttt{NegateMinimize}, since it does not distort the distance
relationships between criteria and therefore works better with
\textsc{topsis}. The data ranks nine major cryptocurrencies (ADA, BNB,
BTC, DOGE, ETH, LINK, LTC, XLM, XRP) using six criteria derived from
historical price and volume series that capture average returns and
their associated risk (variability of returns, of trading volume, and
of trend fit) over two overlapping moving-window sizes, 7 and 15 days,
yielding two decision matrices for the same set of alternatives. We
apply \texttt{RankInvariantChecker} (\textsc{rrt}1) and
\texttt{RankTransitivityChecker} (\textsc{rrt}2/\textsc{rrt}3) to both window sizes,
allowing us to assess not only overall robustness but also its
sensitivity to this experimental design choice.

\begin{figure}[t]
\centering
\includegraphics[width=\textwidth]{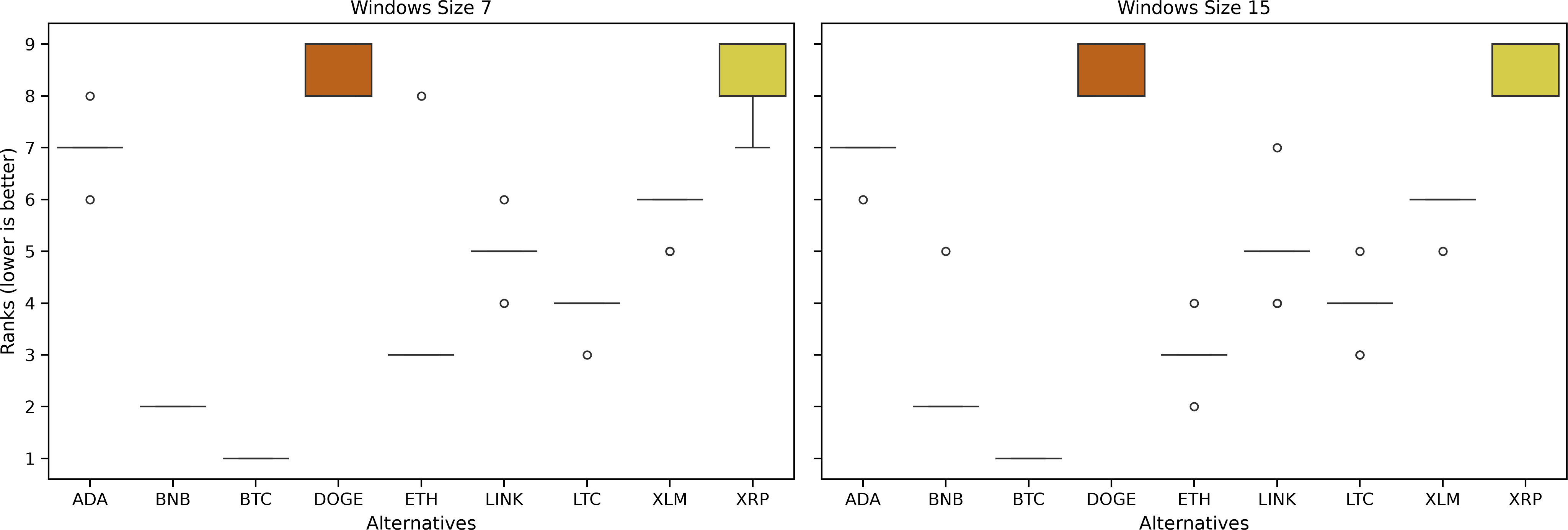}
\caption{Distribution of ranks taken by each alternative across all
\textsc{rrt}1 mutation scenarios, for comparison window sizes of 7 (left) and 15
(right) days. BTC remains pinned at rank 1 in both windows, while the
bottom of the ranking (DOGE, XRP) shows the widest spread, indicating
that instability under degradation concentrates among the
closely-clustered worst alternatives rather than the clearly-separated
best one.}
\label{fig:case1-rrt1-box}
\end{figure}

\textsc{rrt}1 reports a perfect Optimal Preservation Rate ($\text{OPR}=1.000$)
and zero Rank Displacement ($\text{RD}=0.000$) for both window sizes:
Bitcoin (BTC), the winner of the reference ranking, never loses its
top position under any of the controlled degradations. Figure~\ref{fig:case1-rrt1-box}
looks beyond the winner, showing the full distribution of ranks taken
by every alternative across all mutation scenarios.

\begin{figure}[t]
\centering
\includegraphics[width=\textwidth]{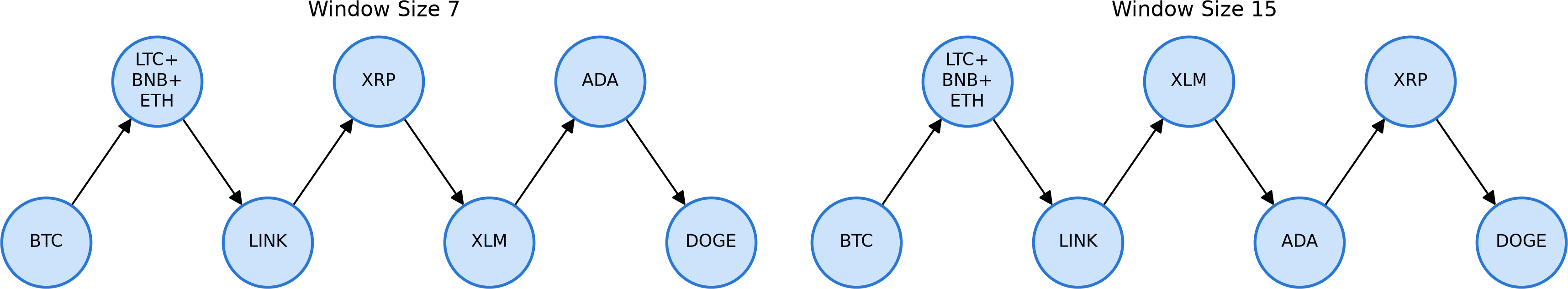}
\caption{Condensed, transitively reduced dominance DAG recovered from
the pairwise comparisons for window sizes of 7 (left) and 15 (right)
days. In both cases, BNB, LTC, and ETH collapse into a single
supernode, the real 3-cycle detected by \textsc{rrt}2, while every other
alternative remains a singleton, preserving a strict chain from BTC
down to DOGE.}
\label{fig:case1-dag}
\end{figure}

\textsc{rrt}2/\textsc{rrt}3, by contrast, uncover an issue invisible to \textsc{rrt}1: both window
sizes fail the transitivity test, with an identical violation rate of
0.033 (one out of the 30 theoretically possible 3-cycles for nine
alternatives), corresponding to a single intransitive triple
$\text{BNB} \succ \text{LTC} \succ \text{ETH} \succ \text{BNB}$ among
the three alternatives immediately below BTC in the reference ranking.
Because \textsc{rrt}2 fails, \textsc{rrt}3 fails as well by construction. Figure~\ref{fig:case1-dag}
shows the condensed dominance DAG obtained from this real pairwise
comparison graph, applying the condensation procedure to the
full \textsc{topsis} pipeline.

Taken together, these results show that a well-behaved, real-world
\mcda pipeline can be fully stable under \textsc{rrt}1 while still violating
transitivity under \textsc{rrt}2/\textsc{rrt}3, confirming that the two families of
tests probe genuinely distinct notions of robustness: stability under
perturbation of the original problem, and internal consistency under
its pairwise decomposition. The fact that the same intransitive triple
and the same violation rate appear under both comparison windows
further suggests that, at least in this case, the inconsistency is a
structural property of the aggregation method on these alternatives
rather than an artifact of the chosen window size.

\subsection{Case 2: Prevalence of Rank Reversal Across Published \mcdm Examples}
\label{sec:case2-literature-audit}

\begin{figure}[t]
\centering
\includegraphics[width=0.7\textwidth]{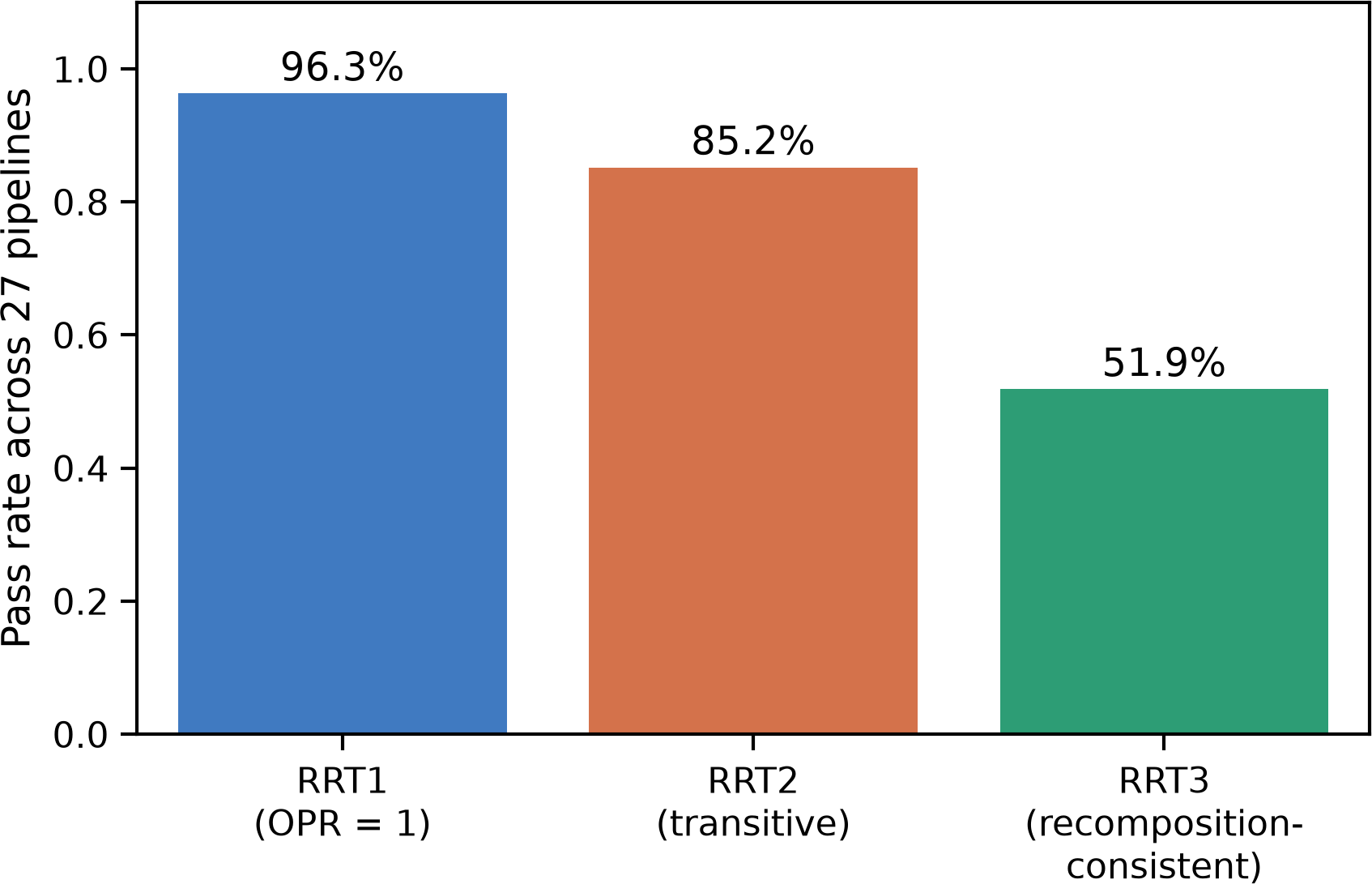}
\caption{Pass rate of \textsc{rrt}1, \textsc{rrt}2, and \textsc{rrt}3 across the 27 published
\mcdm pipeline/dataset combinations audited in Case 2. \textsc{rrt}1 passes
almost universally, \textsc{rrt}2 fails for a minority of pipelines, and \textsc{rrt}3,
the strictest criterion, passes for barely half of the benchmark.}
\label{fig:case2-pass-rates}
\end{figure}

The second case study shifts from depth to breadth. Instead of a
single problem, we reproduce, pipeline for pipeline and matrix for
matrix, 27 worked examples drawn from 20 published \mcdm methods
(including \textsc{codas}~\citep{ghorabaee2016new}, \textsc{copras}~\citep{zavadskas1996new}, \textsc{edas}~\citep{keshavarz2015multi},
\textsc{ervd}~\citep{shyur2015multiple}, \textsc{mabac}~\citep{pamucar20153016}, four \textsc{moora} variants~\citep{brauers2006moora},
\textsc{ocra}~\citep{parkan1994ocra}, \textsc{probid}/Simplified~\textsc{probid}~\citep{wang2021preference}, \textsc{ram}~\citep{SOTOUDEHANVARI2023138695},
\textsc{spotis}~\citep{spotisCometRafsiRRR2021}, \textsc{topsis}, \textsc{vikor}~\citep{opricovic2004compromise}, and
\textsc{waspas}~\citep{zavadskas2012optimization}),
each evaluated with its own originally published
preprocessing (scalers, objective inverters, and method-specific
parameters). In every case we follow the source paper as closely as
possible, reproducing its stated pipeline step for step; when a paper
left some intermediate stage underspecified, we reconstructed that
stage by working backward from the numerical results it reported, so
that our reproduction contemplates not only its inputs but also its
published outputs. Because these are the very examples the original authors
selected to showcase their methods, any rank reversal detected here
cannot be dismissed as an artificial or adversarial construction: it is
already present in the literature's own reference cases. We apply \textsc{rrt}1
and \textsc{rrt}2/\textsc{rrt}3 uniformly to all 27 pipelines and summarize the fraction
that pass each criterion in Figure~\ref{fig:case2-pass-rates}.

\textsc{rrt}1 passes for 96.3\% (26/27) of pipelines, confirming that top-choice
stability under controlled degradation is the common case in the
literature. \textsc{rrt}2 passes for 85.2\% (23/27); the remaining four
pipelines, concentrated around methods built on reference regions or
boundary distances computed from the full alternative set, exhibit
genuine intransitive triples. The gap widens sharply at \textsc{rrt}3, which
passes for only 51.9\% (14/27) of pipelines. Most of this additional
failure comes from pipelines that pass \textsc{rrt}2 (i.e., their pairwise
tournament is perfectly transitive) yet still fail \textsc{rrt}3: this pattern
arises because many \mcdm aggregators compute normalization constants
or ideal/anti-ideal reference points from every alternative present in
the matrix, so scoring alternatives ``all together'' can silently
disagree with scoring them ``one pair at a time,'' a violation of
independence of irrelevant alternatives that \textsc{rrt}2 cannot detect by
construction. This large-scale audit reinforces, on real reference
cases rather than synthetic examples, the central motivation of this
work: transitivity alone is a necessary but insufficient condition for
the kind of decomposition-consistency that \textsc{rrt}3 specifically targets.

It is also worth noting that an audit of this scale would have required, absent a general-purpose, method-agnostic implementation of RRT1-RRT3, a bespoke, manual reimplementation of the degradation, pairwise decomposition, and recomposition logic for each of the 27 pipelines individually. The tooling presented in this paper is what makes this kind of comparative, corpus-wide analysis routine rather than a one-off, per-method exercise.

% ====================================================
\section{Summary and conclusions}
\label{section:summary}
This paper closes a 17-year implementation gap: we operationalized \citet{wang2008ranking}'s three theoretical criteria into concrete, open-source algorithmic tests, built on \skc's \texttt{RankResult} and \texttt{RanksComparator} data structures.

\textsc{rrt}1 evaluates optimal-alternative stability through controlled degradation, with graceful handling of pipelines that eliminate alternatives. \textsc{rrt}2 and \textsc{rrt}3, computed together, assess ranking consistency via a dominance graph built from pairwise comparisons: \textsc{rrt}2 detects transitivity violations through a violation-rate metric, while \textsc{rrt}3 compares the original ranking with those recomposed from that graph after exact condensation and transitive reduction, offering two recomposition strategies (\texttt{generations}, reporting ties directly, and \texttt{cycle\_permutations}, enumerating every plausible strict ordering). Both are available in the \texttt{RankInvariantChecker} and \texttt{RankTransitivityChecker} classes within \skc's rank reversal module\footnote{Available at \url{https://github.com/quatrope/scikit-criteria/blob/dev/skcriteria/ranksrev/}}.

Limitations include a degradation strategy that may not capture all real-world perturbations and an $O(n^2)$ cost for \textsc{rrt}2/\textsc{rrt}3 that may limit scalability beyond 50 alternatives; both point to future work, along with extending the pairwise decomposition to group decision-making and scaling the Case 2 audit to a larger corpus of published \mcdm examples.
The framework is available starting from \skc version 0.10 through \texttt{pip install scikit-criteria}.

\small
\section*{Acknowledgements}

This work was partially supported by the Consejo Nacional de Investigaciones Científicas y Técnicas (CONICET) and Comisión Nacional de Actividades Espaciales (CONAE). B.A., G.I.D.N., S.A.R., G.G. were supported by CONICET fellowships, while P.P. was supported by a fellowship from CONICET and CONAE.

The project was developed within the framework of the PhD course \textit{Advanced Software Design Techniques}, at the Facultad de Matemática, Astronomía, Física y Computación (FAMAF).

\section*{Declaration of generative AI and AI-assisted technologies in the writing process}
 During the preparation of this work the author(s) used OpenAI-ChatGPT and Anthropic-Claude in order to improve the clarity and correctness of the text. After using this tool/service, the author(s) reviewed and edited the content as needed and take(s) full responsibility for the content of the published article.

% \appendix

% \section{Appendix title 1}
% %% \label{}

% \section{Appendix title 2}
% %% \label{}

\bibliographystyle{apalike}
\bibliography{example}



%% For example:

%% \bibitem[Aladro et al.(2015)]{Aladro15} Aladro, R., Martín, S., Riquelme, D., et al. 2015, \aas, 579, A101
%%\end{thebibliography}

\end{document}

%% file: algorithm/rrt1.tex
\begin{algorithmic}[1]
\Require Decision Matrix $D$, \mcda Method $M$, Repetitions $R$
\Ensure RanksComparator with $(|Alternatives|-1) \times R + 1$ rankings
\State Baseline: Evaluate $D$ with $M \rightarrow Ranking_0$ (reference)
\For{each repetition $r \in [1, R]$}
    \For{each suboptimal alternative $A_i$}
        \State Generate degraded alternative $A'_i$
        \State Create modified matrix $D'_i = D.\text{replace}(A_i, A'_i)$
        \State Evaluate $D'_i$ with $M \rightarrow Ranking_{i,r}$
        \State Report mutation details in $Ranking_{i,r}.\text{extra\_}$
    \EndFor
\EndFor
\State \Return RanksComparator($[Ranking_0, \{Ranking_{i,r}\}]$)
\end{algorithmic}

%% file: algorithm/graceful_degradation.tex
\begin{algorithmic}[1]
\Require Current alternatives in ranking: $alternatives$, Original alternatives from decision matrix: $full\_alternatives$, Boolean flag for missing alternatives policy: $allow\_missing$, Current ranking values: $values$
\Ensure Updated $alternatives$ and $values$ with missing alternatives handled
\State $missing\_alts \leftarrow full\_alternatives \setminus alternatives$
\If{$|missing\_alts| > 0$}
    \If{$allow\_missing = \text{True}$}
        \State $max\_rank \leftarrow \max(values)$
        \State $fill\_values \leftarrow [max\_rank + 1] \times |missing\_alts|$
        \State $alternatives \leftarrow alternatives \cup missing\_alts$
        \State $values \leftarrow values \cup fill\_values$ \Comment{Assign worst rank to filtered alternatives}
    \Else
        \State \textbf{raise} Error("Pipeline eliminated alternatives")
    \EndIf
\EndIf
\State \Return $alternatives$, $values$
\end{algorithmic}

%% file: algorithm/rrt2.tex
\begin{algorithmic}[1]
\Require Decision Matrix $D$, \mcda Method $M$
\Ensure Test result, transitivity break list, transitivity break rate
\State Baseline: Evaluate $D$ with $M \rightarrow Ranking_0$ (reference)
\State $pairs \gets \text{all\_combinations}(Ranking_0.\text{alternatives}, 2)$
\For{each pair $(A_i, A_j) \in pairs$} \Comment{Parallel execution}
    \State Create sub-matrix $D_{ij} = D.\text{loc}[A_i, A_j]$
    \State Evaluate $D_{ij}$ with $M \rightarrow Ranking_{ij}$
\EndFor
\State $graph \gets \text{DiGraph}()$
\For{each $Ranking_{ij}$}
    \State $winner \gets \arg\min(Ranking_{ij}.\text{rank})$
    \State $loser \gets \arg\max(Ranking_{ij}.\text{rank})$
    \State $graph.\text{add\_edge}(winner, loser)$
\EndFor
\State $trans\_break \gets \text{simple\_cycles}(graph)$
\State $test\_criterion\_2 \gets (|trans\_break| = 0)$
\State \Return $(test\_criterion\_2, graph, trans\_break)$
\end{algorithmic}